\documentclass{article}

\usepackage{microtype}
\usepackage{graphicx}
\usepackage{subcaption}
\usepackage{booktabs} 

\usepackage{hyperref}

\usepackage{pifont}
\usepackage[preprint]{icml2026}

\usepackage{amsmath}
\usepackage{amssymb}
\usepackage{mathtools}
\usepackage{amsthm}

\usepackage{booktabs}       
\usepackage{tabularx}
\usepackage{multirow}
\usepackage{longtable}

\usepackage{bm}

\usepackage[capitalize,noabbrev]{cleveref}

\theoremstyle{plain}
\newtheorem{theorem}{Theorem}[section]

\newtheorem{lemma}[theorem]{Lemma}

\theoremstyle{definition}

\theoremstyle{remark}

\usepackage[textsize=tiny]{todonotes}

\icmltitlerunning{Submission and Formatting Instructions for ICML 2026}

\begin{document}

\twocolumn[
\icmltitle{Winner-Take-All Spiking Transformer for Language Modeling}


\icmlsetsymbol{equal}{*}

\begin{icmlauthorlist}
\icmlauthor{Chenlin Zhou}{111,222}
\icmlauthor{Sihang Guo}{333}
\icmlauthor{Jiaqi Wang}{222,333}
\icmlauthor{Dongyang Ma}{444}
\icmlauthor{Kaiwei Che}{111,222}
\icmlauthor{Baiyu Chen}{222} \\
\icmlauthor{Qingyan Meng}{222} 
\icmlauthor{Zhengyu Ma}{222}
\icmlauthor{Yonghong Tian}{111,222,444}
\end{icmlauthorlist}

\icmlaffiliation{111}{School of Electronic and Computer Engineering, Shenzhen Graduate School, Peking University}
\icmlaffiliation{222}{Pengcheng Laboratory}
\icmlaffiliation{333}{Harbin Institute of Technology}
\icmlaffiliation{444}{School of Computer Science, Peking University}

\icmlcorrespondingauthor{Zhengyu Ma}{mazhy@pcl.ac.cn}
\icmlcorrespondingauthor{Yonghong Tian}{yhtian@pku.edu.cn}

\icmlkeywords{Machine Learning, ICML}

\vskip 0.3in
]



\printAffiliationsAndNotice{\icmlEqualContribution} 

\begin{abstract}
Spiking Transformers, which combine the scalability of Transformers with the sparse, energy-efficient property of Spiking Neural Networks (SNNs), have achieved impressive results in neuromorphic and vision tasks and attracted increasing attention. 
However, existing directly trained spiking transformers primarily focus on vision tasks. For language modeling with spiking transformer, convergence relies heavily on softmax-based spiking self-attention, which incurs high energy costs and poses challenges for neuromorphic deployment.
To address this issue, we introduce Winner-Take-All (WTA) mechanisms into spiking transformers and propose two novel softmax-free, spike-driven self-attention modules: WTA Spiking Self-Attention (WSSA) and Causal WTA Spiking Self-Attention (CWSSA). Based on them, we design WTA-based Encoder-only Spiking Transformer (WE-Spikingformer) for masked language modeling and WTA-based Decoder-only Spiking Transformer (WD-Spikingformer) for causal language modeling, systematically exploring softmax-free, spiking-driven Transformer architectures trained end-to-end for natural language processing tasks.
Extensive experiments on 16 datasets spanning natural language understanding, question-answering tasks, and commonsense reasoning tasks validate the effectiveness of our approach and highlight the promise of spiking transformers for general language modeling and energy-efficient artificial intelligence.

\end{abstract}

\section{Introduction}
Spiking Neural Networks (SNNs), regarded as the third generation of neural networks \citep{maass1997networks}, offer high biological plausibility and energy efficiency through their event-driven dynamics, making them strong contenders to Artificial Neural Networks (ANNs) \citep{roy2019towards}.
By transmitting information through binary spikes, SNNs replace traditional high-power multiply-accumulate (MAC) operations with low-power accumulate (AC) operations, thereby achieving substantial energy savings.

Spiking transformers, which combine the architectural strengths of Transformers with the event-driven, sparse, and energy-efficient properties of spiking neural networks, have achieved great progress on both neuromorphic datasets and large-scale vision datasets \citep{zhou2023spikformer,zhou2023spikingformer,yao2023spike,yao2024spike,zhou2024qkformer,yao2025scaling} and have attracted significant attention.
They achieve performance comparable to that of ANN counterparts while offering superior energy efficiency.
Current directly trained spiking transformers have primarily targeted computer vision tasks \citep{zhou2023spikformer,zhou2023spikingformer,yao2023spike,yao2024spike,zhou2024qkformer,yao2025scaling} with encoder-only architectures. 
Spiking transformers with softmax-free self-attention perform well on visual tasks because image patches are locally correlated and highly redundant. Repeated low-level features, such as edges and textures, allow softmax-free attention to achieve competitive performance. 

By contrast, as the extension to spiking LLM is a growing trend, language signals are sparse and heavily depend on long-range dependencies, making the above softmax-free spiking transformers far more difficult to design for language modeling. For example,
SpikeBert \citep{lv2023spikebert} is an early attempt in this direction. As a softmax-free spiking transformer, SpikeBert adapts Spikformer \citep{zhou2023spikformer} for language tasks by proposing a two-stage knowledge distillation training method, and achieves 59.7$\%$ accuracy on the natural language understanding benchmark (GLUE dev datasets) \citep{wang2018glue}. The substantial performance gap between SpikeBert and BERT \citep{devlin2019bert} (19.9\%) highlights the difficulty of directly applying spiking self-attention from a vision-based spiking transformer (spikformer) to language modeling.
%
More recently, SpikeLM \citep{xing2024spikelm} and SpikeLLM \citep{xing2024spikellm} have achieved competitive results in language modeling; however, they do not fully adhere to a spike-driven design. SpikeLM retains non-spiking GeLU activations \citep{hendrycks2016gaussian} in the MLP blocks, while SpikeLLM preserves SiLU activations in its MLP blocks. More importantly, both SpikeLM and SpikeLLM are softmax-based spiking transformers. The softmax with complex exponential and division operations brings huge challenges to energy consumption and neuromorphic deployment \citep{zhou2023spikformer, zhou2023spikingformer}.

%

To address these limitations, we explore softmax-free spike-driven transformers, tailor-made for language modeling.  
Firstly, we introduce a brain-inspired mechanism: Winner-Take-All (WTA) for spiking transformers to replace the softmax operation in language modeling. 
Winner-take-all mimics biological lateral inhibition to enforce sparsity and focus attention on the most relevant tokens, serving as an extremely sparse alternative to softmax, which makes it particularly suitable for spiking language modeling.
Through incorporating the WTA biological mechanisms, we developed two kinds of softmax-free, spike-driven transformers without floating-point multiplications: WTA-based Encoder-only SpikingFormer for masked language modeling and WTA-based Decoder-only SpikingFormer for causal language modeling. These architectures expand the spiking transformer family and advance both neuromorphic intelligence and energy-efficient artificial intelligence. Our main contributions are as follows :

1) Leveraging spike-driven and Winner-Take-All (WTA) biological mechanisms, we proposed two novel spike-driven self-attention for language modeling: WTA Spiking Self-Attention (\textbf{WSSA}) and Causal WTA Spiking Self-Attention (\textbf{CWSSA}). 
In contrast to conventional self-attention that relies on costly floating-point multiplications and softmax, our approach operates solely with sparse additions, eliminating the need for expensive computation.

2) We develop a WTA-based Encoder-only Spiking Transformer (\textbf{WE-Spikingformer}) with WSSA for masked language modeling, and a WTA-based Decoder-only Spiking Transformer (\textbf{WD-Spikingformer}) with CWSSA for causal language modeling, systematically exploring direct-training-based softmax-free spike-driven transformers in language modeling.

3) We evaluate our models on 16 datasets covering natural language understanding, question-answering, and commonsense reasoning tasks. Extensive experiments demonstrate the promise of directly trained spiking transformers for general language modeling and energy-efficient artificial intelligence.

\section{Related Works}
\subsection{Spiking Transformers in Vision Tasks}
Spiking transformers with encoder-only architectures have been widely adopted in visual tasks \citep{zhou2023spikformer, zhou2023spikingformer,yao2023spike,yao2024spike,zhou2024qkformer, yao2025scaling}.
Spikformer \citep{zhou2023spikformer} introduced Spiking Self-Attention (SSA), which replaces softmax with sparse spike-form Query, Key, and Value, achieving 74.81$\%$ accuracy on ImageNet-1k with only four time steps—the first strong evidence of the potential of transformer-based SNNs. Building on it, Spikingformer \citep{zhou2023spikingformer} employed a pre-activation shortcut to eliminate floating-point multiplications and reduce firing rates, further improving accuracy to 75.85$\%$. Spike-driven Transformer \citep{yao2023spike} proposed Spike-Driven Self-Attention (SDSA), which relies solely on masking and addition, reaching 77.07$\%$ on ImageNet-1k with significantly lower computational cost. Recently, hierarchical visual spiking transformers \citep{yao2024spike,zhou2024qkformer, yao2025scaling} has achieved a performance of over 80$\%$ on ImageNet while maintaining high energy efficiency.
MaxFormer \cite{fangspiking} and SpikePool \cite{lee2025spikepool} observe that spiking neurons tend to suppress certain frequency components during information transmission. To compensate for this effect, they introduce max-pooling operations at different positions in Spiking Transformers to recover informative signals: MaxFormer applies max pooling in the patch embedding stage to restore high-frequency components, while SpikePool replaces spike-based self-attention with max-pooling attention to achieve a selective band-pass filtering effect.
These methods are primarily designed for vision tasks and are hard to generalize to large-scale, data-driven scenarios such as language modeling. In contrast, our work targets language modeling and introduces Hard Winner-Take-All directly on the attention map within the self-attention layer, acting as a one-hot encoding mechanism to produce extremely sparse yet scalable attention.

\subsection{Spiking Transformers in Language Tasks}
SpikeBert \citep{lv2023spikebert} is a softmax-free spiking transformer, which improves Spikformer \citep{zhou2023spikformer} to process language tasks and propose a two-stage knowledge distillation method for training it. The two-stage knowledge distillation method combines pretraining by distilling knowledge from BERT with a large collection of unlabelled texts and fine-tuning with task-specific instances via knowledge distillation from the BERT fine-tuned on the same training examples, and achieves 59.7$\%$ accuracy on the GLUE development benchmark \citep{wang2018glue}, which is nearly 20$\%$ lower than competitive language models, indicating that directly applying vision-oriented spiking self-attention to language tasks results in suboptimal performance.
SpikeGPT \citep{zhu2023spikegpt} is an RWKV \citep{peng2023rwkv} architecture-based spiking language model and retains the exponential and division operations similar to the softmax operation.
SpikeLM \citep{xing2024spikelm} proposes a spike formulation with bi-directional ternary firing for language modeling and achieves promising results in language modeling; however, it largely preserves key components of vanilla attention, including floating-point matrix multiplications, softmax operations, and non-spiking activations (e.g., GeLU \citep{hendrycks2016gaussian}) in the MLP blocks.
Similarly, SpikeLLM \citep{xing2024spikellm} proposes generalized integrate-and-fire (GIF) neurons together with an optimal brain spiking framework for spiking transformers. However, SpikeLLM largely inherits from LLaMA \citep{touvron2023llama} and retains many nonlinear components from LLaMA, such as softmax operations in attention map computation, non-spiking SiLU \citep{hendrycks2016gaussian} activations in the MLP blocks, and rotary positional embeddings \citep{su2024roformer} with non-spike-driven computations in the query and key projections.
By contrast, our proposed WE-Spikingformer and WD-Spikingformer eliminate these non-spiking components to realize a spike-driven transformer architecture. In short, this work focuses on designing softmax-free, energy-efficient spike-driven transformers for language modeling.

%
%


\section{Method}


\subsection{Spiking Neuron Model}
In this section, to investigate more expressive yet efficient spiking neurons for language modeling, we also explore two variants of the Leaky Integrate-and-Fire (LIF) neuron model \citep{maass1997networks} for direct training in language modeling tasks: T-LIF \citep{xing2024spikelm} and NI-LIF \citep{lei2025spike2former}.

\textbf{Ternary-spiking-based Leaky Integrate-and-Fire (T-LIF) model.}
Compared to the original LIF neuron, the ternary spiking in SpikeLM \citep{xing2024spikelm} extends binary spikes $\{0, 1\}$ to ternary values $\{-\alpha, 0, \alpha\}$ based on membrane potential intensity. The dynamics of the T-LIF model  are formulated as:
\begin{align}
& U[t]=H[t-1]+X[t], \\
& {S}[t]= \begin{cases}-1\cdot{\alpha}[t], & \text { if } {U}[t]<-{\alpha}[t], \\ 0\cdot{\alpha}[t], & \text { if } {U}[t] \in(-{\alpha}[t],+{\alpha}[t]), \\ +1 \cdot{\alpha}[t], & \text { if } {U}[t]>+{\alpha}[t],\end{cases} \\
& H[t]=V_{\text {reset }} |b|+(\beta U[t])({1}-|b|),
\end{align}
where $X[t]$ is the input current at time step $t$, ${S}[t]=b\cdot\alpha[t]$ and $b \in \{-1,0,1\}$. $U[t]$ represents the membrane potential after the triggered event, which will decay directly to $H[t]$ if no spike is generated (where $\beta < 1$ is the decay factor) and otherwise equals to the reset potential $V_{reset}$.

\textbf{Normalized Integer Leaky Integrate-and-Fire (NI-LIF) model.} 
This neuron adopts integer-training and spike-inference way \citep{luo2024integer}, the dynamics of the NI-LIF model  are formulated as follows:
\begin{align}
&U[t]=H[t-1]+X[t], \\
&S[t]=\operatorname{c l i p}(\operatorname{round}(U[t]), 0, D) / D, \\
&H[t]=\beta(U[t]-S[t] \times D),
\end{align}
where $\operatorname{clip}(U[t], min, max)$ denotes  the operation of clipping $U[t]$ to $[min, max]$, $D$ indicates the maximum quantized integer value and unfold into $D$ time steps when inference on neuromorphic chips. That is, the total time step $\operatorname{T}$ of the spike sequence is $T*D$, where $T$ is the normalized integer time step. For example, NI-LIF(1$\times$4 ) unfolds into a binary spike sequence of time step $\operatorname{T}$ = 4. This neuronal mechanism allows for faster training compared to T-LIF.

In our experiments, we use NI-LIF neurons by default for faster spiking simulation, and we also compare the performance differences between the two neurons in Section.\ref{NLU}.


\subsection{Winner-Take-All Layer}

\begin{figure*}[!tb]
\centering
\includegraphics[width=0.9\textwidth]{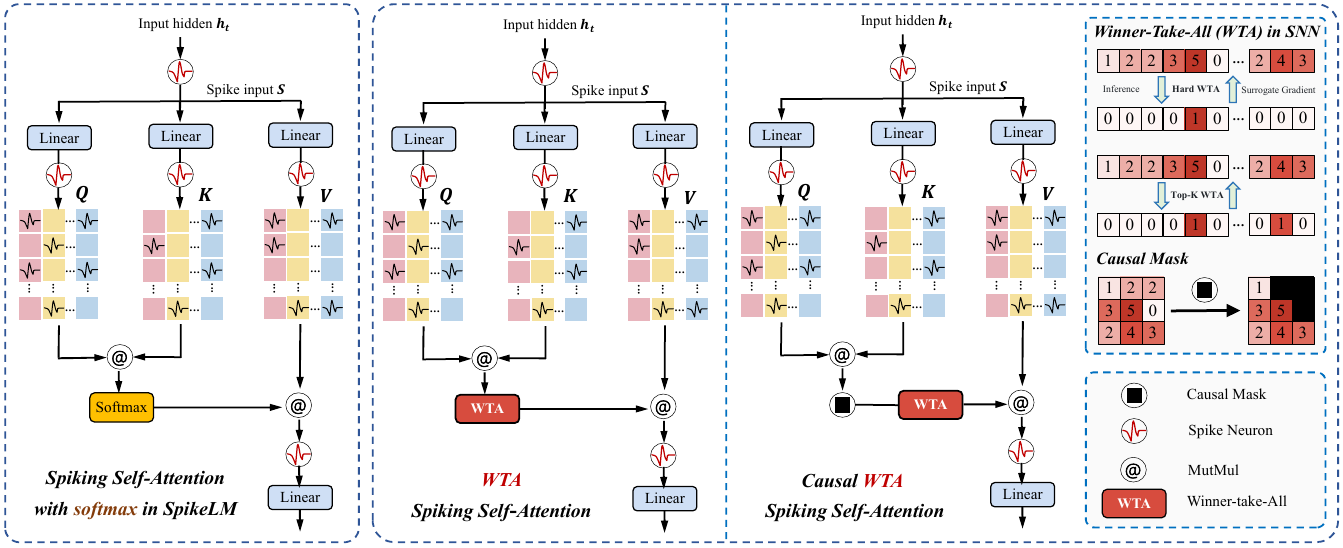}
\caption{Overview of WTA Spiking Self-Attention (WSSA) and Causal WTA Spiking Self-Attention (CWSSA). The left shows the softmax-based spiking self-attention in SpikeLM \citep{xing2024spikelm}. The right shows our WSSA in WE-Spikingformer for masked language modeling and CWSSA in WD-Spikingformer for causal language modeling. 
The symbol “@” denotes matrix multiplication. 
}
\vspace{-3 mm}
\label{Attention}
\end{figure*}

In the visual system, multiple neurons are sensitive to different directions at the same location. The neurons with the strongest response (such as detecting vertical edges) will suppress other direction detectors, leaving only a dominant signal. 
The Winner-Take-All computation \citep{maass2000computational}, inspired by this biological mechanism, is incorporated into spiking neural networks as a biologically plausible strategy for enforcing sparsity.
It simulates the "lateral inhibition" phenomenon in biological nervous systems - that is, the strongest signal inhibits other competing signals, ultimately forming a single dominant response and ensuring that attention remains focused on the most relevant token. 
In this section, we primarily introduce multiple WTA mechanisms and their corresponding training methods.

\textbf{Hard WTA.}
Hard WTA is an sparse neural activation mechanism. In a group of competing units, only the unit with the highest activation value is allowed to retain the original output, and the outputs of all other units are forced to zero. Given an input vector $\operatorname{A}= [a_1, \ldots, a_n]$, the output layer of WTA is $\operatorname{Y}= [y_1, \ldots, y_n]$. Hard WTA can be formulated as follows:
\begin{equation}
y_i= \begin{cases}a_i  & \text { if } i=\arg \max _{j \in ( 1,...,N) } a_j, \\ 0 & \text { otherwise}.\end{cases}
\end{equation}
In SNN, the winning element needs to be binarized, that is, the winner is further defined as: $y_i= 1, \text { if } i=\arg \max _{j \in ( 1,...,N) } a_j$. 
Hard WTA in SNN is an extremely sparse softmax with one-hot encoding. Appendix \ref{Hard_WTA} provides a detailed discussion of the relationship between HardWTA and softmax.

\textbf{Top-K WTA.} As a sparse neural activation mechanism, Top-K WTA selects the K units with the highest scores from a set of input signals, retains only their original values, and sets the outputs of all other units to zero. Top-K WTA allows a limited number of "winners" to coexist, thus achieving a balance between sparsity and expressiveness. Top-K WTA can be formulated as follows:
\begin{equation}
y_i= \begin{cases}a_i & \text { if } a_i \in \operatorname{Top-k}\left(a_1, \ldots, a_n\right), \\ 0 & \text { otherwise}.\end{cases}
\end{equation}
In SNN, the top-k winning elements are further binarized, that is: $y_i= 1, \text { if } a_i \in \operatorname{Top-k}\left(a_1, \ldots, a_n\right)$. 

\textbf{Sparsemax.}
Sparsemax \cite{martins2016softmax} is a differentiable, adaptively sparse neural mechanism that can automatically generate some zero values in the output, which can also be seen as an adaptive Top-K Winner-Take-All mechanism.
That is, only a few significant elements are given non-zero probability, and the remaining elements are precisely truncated to zero, thereby achieving adaptive sparsity. The calculation process of Sparsemax can be described as follows:
\begin{equation}
y_i=\max \left\{a_i-\tau, 0\right\}, \quad i=1, \ldots, N . \label{Sparsemax}
\end{equation}
Eq. \ref{Sparsemax} needs to satisfy $\sum_i y_i=1$, and $\tau$ is the adaptive threshold, whose calculation is shown in Appendix \ref{appendix_sparsemax}.


\begin{figure*}[!tb]
\centering
\includegraphics[width=0.9\textwidth]{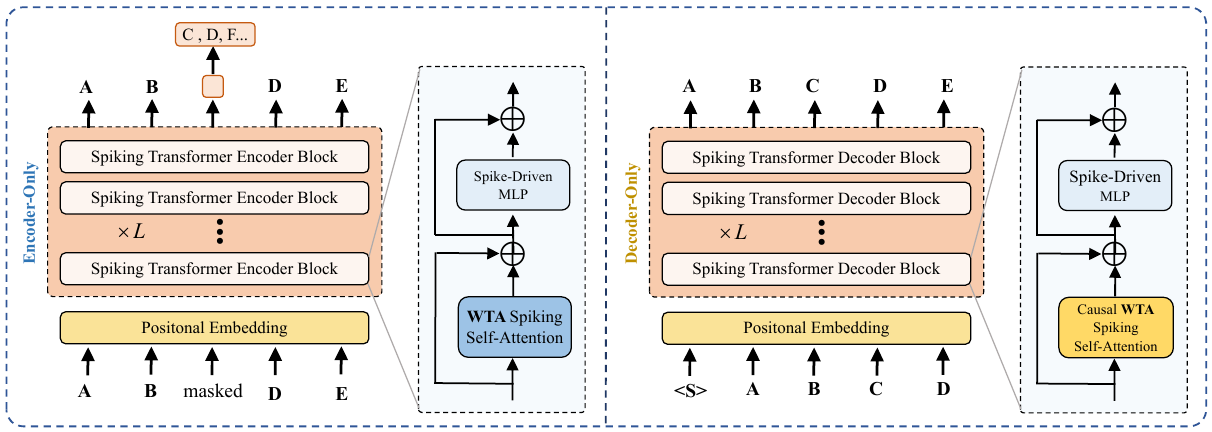}
\caption{The overview of WE-Spikingformer and WD-Spikingformer. The left shows WE-Spikingformer (WTA-based Encoder-only Spiking Transformer) for spike-based masked language modeling. The right shows WD-Spikingformer (WTA-based Decoder-only Spiking Transformer) for spike-based causal language modeling.}
\label{Encoder_Decoder}
\end{figure*}

\textbf{Surrogate Gradient for the WTA Layer.}
The Winner-Take-All mechanism is non-differentiable because it produces discrete, discontinuous outputs with zero gradients almost everywhere and undefined gradients at decision boundaries. 
Therefore, we choose the surrogate gradient for Winner-Take-All during training, which does not affect the high energy efficiency of the model during inference. 
Specifically, we use the gradient of softmax to approximate the "gradient" of the Winner-Take-All layer, which is shown as follows:

\begin{equation}
\frac{\partial y}{\partial a} = \mathrm{softmax}'(a).
\end{equation}
where $\mathrm{softmax}'(.)$ is the derivative of the softmax function.


Winner-Take-All is functionally equivalent to an extremely sparse version of softmax, ensuring that attention remains focused on the most relevant token. It uses the lateral inhibition mechanism in neural dynamics instead of exponential normalization. It is more suitable for spiking language modeling due to its sparse dependencies and lateral inhibition, while also offering energy efficiency advantages.
In our experiments, we exploit Hard WTA by default owing to its superior energy efficiency.

\subsection{WTA Spiking Self-Attention Mechanisms} \label{WSSA_and_CWSSA}

In this part, we propose two self-attention mechanisms based on the Winner-Take-All layer for spiking language modeling: WTA Spiking Self-Attention (WSSA) and Causal WTA Spiking Self-Attention (CWSSA), shown in Figure \ref{Attention}.

\textbf{CWSSA} is a spike-driven self-attention for spike-based causal language modeling with Decoder-only Transformers. CWSSA can be formulated as follows:
\begin{align}
\textbf{X}^{\prime} &= \operatorname{SN}(\textbf{X}), \\
\textbf{Q} &= \operatorname{SN}_{Q}(\operatorname{Linear}_{Q}(\textbf{X}^{\prime})), \\
\textbf{K} &= \operatorname{SN}_{K}(\operatorname{Linear}_{K}(\textbf{X}^{\prime})), \\
\textbf{V} &= \operatorname{SN}_{V}(\operatorname{Linear}_{V}(\textbf{X}^{\prime})), \\
\operatorname{A_{w}}(\textbf{Q}, \textbf{K}) &= \operatorname{WTA}(\mathbb{M}(\textbf{Q} \textbf{K}^{\mathrm{T}} * s )), \\ \label{cwssa}
\operatorname{Out}(\operatorname{A_{w}}, \textbf{V}) &= \operatorname{Linear}(\operatorname{SN}(\operatorname{A_{w}} \textbf{V})),
\end{align}
where $s$ is the scaling factor, same in \citet{zhou2023spikformer},  $\operatorname{A_{w}}$ is the attention weights, and $\operatorname{SN}$ means the spiking neuron. $\mathbb{M}(.)$ means the causal mask. 

\textbf{WSSA} is a spike-driven self-attention for spike-based masked language modeling with Encoder-only Transformers.
The attention mechanism is similar to CWSSA, with the main difference that the causal mask in Eq. \ref{cwssa} is removed:
\begin{equation}
\operatorname{A_{w}}(\textbf{Q}, \textbf{K}) = \operatorname{WTA}((\textbf{Q} \textbf{K}^{\mathrm{T}} * s)).  \label{wssa}
\end{equation}

Compared to the softmax-based SSA in SpikeLM \citep{xing2024spikelm}, both WSSA and CWSSA are spike-driven and more energy-efficient, avoiding the costly exponential and division operations required by softmax.


\subsection{Language Modeling}

Based on WSSA and CWSSA in Section~\ref{WSSA_and_CWSSA}, we build two spiking transformers for spike-based language modeling: WTA-based Spiking Encoder (WE-Spikingformer) and WTA-based Spiking Decoder (WD-Spikingformer) in this part.

\textbf{Model Architecture.} The overview of WE-Spikingformer and WD-Spikingformer is shown in Figure ~\ref{Encoder_Decoder}. The decoder block in WD-Spikingformer can be formulated as follows:
\begin{align}
&X_l^{\prime}=\operatorname{CWSSA}\left(X_{l-1}\right)+X_{l-1},  \quad X_l^{\prime} \in \mathbb{R}^{T \times d}, l=1 \ldots L ,\label{CWSSA} \\
&X_l=\operatorname{SMLP}\left(X_l^{\prime}\right)+X_l^{\prime},  \quad X_l \in \mathbb{R}^{T \times d}, l=1 \ldots L,
\end{align}
where $\operatorname{SMLP}$ means Spike-driven Multi-Layer Perceptron.
Unlike SpikeLM and SpikeLLM, which retain non-spiking activation functions such as GeLU or SiLU, our 
$\operatorname{SMLP}$ is fully spiking and implemented as a sequential stack of \{SN-Linear-SN-Linear\} layers.
The encoder block in WE-Spikingformer is similar to the decoder one in WD-Spikingformer, with the main difference that CWSSA in Eq.~\ref{CWSSA} is replaced by WSSA in Eq.~\ref{WSSA}:
\begin{equation}
X_l^{\prime}=\operatorname{WSSA}\left(X_{l-1}\right)+X_{l-1},  \quad X_l^{\prime} \in \mathbb{R}^{T \times d}, l=1 \ldots L .\label{WSSA} 
\end{equation}

\textbf{Masked Language Modeling.}
Given a sequence of tokens $\operatorname{x}= (x_1, x_2, \ldots, x_T)$, randomly select a portion of the locations $M \subseteq\{1,2, \ldots, T\}$ for masking. The masked sequence is obtained as $\tilde{x}=\left(\tilde{x}_1, \tilde{x}_2, \ldots, \tilde{x}_T\right)$. The goal of Masked Language Modeling (MLM) \citep{devlin2019bert} is to let the model predict the masked tokens based on the unmasked tokens, which can be formulated as follows:
\begin{equation}
p\left(x_M \mid \tilde{x}_{\backslash M}\right),
\end{equation}
where $x_M=\left\{x_t: t \in M\right\}$ denotes the masked token, $\tilde{x}_{\backslash M}$ denotes the visible tokens that are retained.
Thus, the loss function of MLM can be formulated as follows:
\begin{equation}
\mathcal{L}_{\mathrm{MLM}}=-\sum_{t \in M} \log p_\theta\left(x_t \mid \tilde{x}\right).
\end{equation}
In our work, we apply WE-Spikingformer to masked language modeling for pretraining.

\begin{table*}[!tb]
\begin{center}
\begin{small}
\centering
\caption{The results on the  Natural Language Understanding task (GLUE datasets). "Avg." denotes "Average Accuracy ". "SD" denotes "Spike-Driven". The results of LIF-BERT, PSN-BERT, and SpikeLM  are reported in \citet{xing2024spikelm}. WE-Spikingformer uses NI-LIF neuron, while WE-Spikingformer$^\dagger$ adopts the T-LIF neuron. Our model achieves state-of-the-art performance on the GLUE datasets among softmax-free spike-driven transformer models.
}
\label{tab_glue}
\setlength{\tabcolsep}{1.7mm}
\fontsize{9.3pt}{\baselineskip}\selectfont
\begin{tabular}{lcp{0.4cm}<{\centering}|cccccccc|p{1.2cm}<{\centering}}
\toprule
{Model}  & {SD} & {T}  & {MNLI\_{mm}} & {QQP} & {QNLI} & {SST-2} & {CoLA} & {STS-B} & {MRPC} & {RTE} & {Avg. (\%)} \\ 
\midrule
BERT$_\texttt{base}$ \citep{devlin2019bert}  &\ding{55} & --   & 83.4 & 71.2 & 90.5 & 93.5 & 52.1 & 85.8 & 88.9 & 66.4 & 79.6 \\
Q2BERT \citep{zhang2020ternarybert}  &\ding{55}  & --   & 47.3 & 67.0 & 61.3 & 80.6 & 0.0  & 4.7  & 81.2 & 52.7 & 49.1 \\
BiBERT \citep{qin2022bibert}  &\ding{55} & --   & 67.5 & 84.8 & 72.6 & 88.7 & 25.4  & 33.6  & 72.5 & 57.4 & 63.2 \\

SpikeLM \citep{xing2024spikelm} &\ding{55} & 4   & 77.2 & 83.9 & 85.3 & 87.0 & 38.8 &  84.9 &  85.7 & 69.0 & 76.5 \\
\midrule
LIF-BERT \citep{xing2024spikelm} &\ding{51}  & 4   & 35.2   & 0 & 50.5 & 50.9 & 0 & 0 & 81.2 & 52.7 & 34.6 \\
PSN-BERT \citep{xing2024spikelm} &\ding{51}  & 4   & 35.2   & 0 & 50.5 & 50.9 & 0 & 6.8 & 81.2 & 52.7 & 34.7 \\
SpikeBERT \citep{lv2023spikebert} &\ding{51}   & 4    & 71.0 & 68.2 & 66.4 & 85.4 & 16.9 & 18.7 & 82.0 & 57.5 & 59.7 \\
{WE-Spikingformer (ours)} &\ding{51}   & 4    & 73.4 & 83.8 & 78.3 & 85.6 & 23.7 & 56.0 & 76.4 & 48.4 & {65.7} \\
{WE-Spikingformer$^\dagger$ (ours)} &\ding{51}   & 4    & 70.1 & 85.1 & 77.5 & 89.0 & 27.9 & 42.8 & 81.6 & 55.9 & \textbf{66.3} \\
\bottomrule
\end{tabular}
\end{small}
\end{center}
\end{table*}

\textbf{Causal Language Modeling.}
Causal Language Modeling (CLM) \citep{radford2018improving, ouyang2022training} is the most widely used language modeling method. Its core idea is autoregressive generation: when predicting the next word, only the context before the current position is allowed to be used, and future words cannot be accessed.
Given a sequence of tokens $\operatorname{x}= (x_1, x_2, \ldots, x_T)$. 
The training strategy of CLM can be formulated as follows:
\begin{equation}
p(x)=\prod_{t=1}^T p\left(x_t \mid x_{<t}\right),
\end{equation}
where $x_t$ represents the $t$-th token, $x_{<t}=(x_1, x_2, \ldots, x_{t-1})$ represents all previous tokens. The optimization goal of CLM is to maximize its conditional probability decomposition. 
In network modeling, the causal mask is used to ensure that the model only focuses on the previous tokens when calculating attention. The loss function of CLM can be formulated as follows:
\begin{equation}
\mathcal{L}_{\mathrm{CLM}}=-\sum_{t=1}^T \log p_\theta\left(x_t \mid x_{<t}\right).
\end{equation}
In our work, we apply WD-Spikingformer for causal language modeling for pretraining.

\section{Experiments}
In this section, we conduct experiments across 16 datasets spanning natural language understanding tasks, question-answering tasks, and commonsense reasoning tasks to evaluate our Winner-Take-All spiking transformer in language modeling.

\subsection{Natural Language Understanding} \label{NLU}
We evaluate WE-Spikingformer on the standard GLUE (General Language Understanding Evaluation) benchmark \citep{wang2018glue}, which is a widely adopted collection of datasets designed to evaluate and advance natural language understanding capabilities of machine learning models. 
GLUE contains 8 subsets for classification and regression, including single-sentence classification (CoLA, SST-2), pairwise sentence comparison (MPRC, QQP, RTE), sentence similarity (STS-B), and natural language inference (MNLI, QNLI). 

We pretrain WE-Spikingformer and WE-Spikingformer$^\dagger$ on Wikipedia-English \citep{devlin2019bert} using masked language modeling \citep{devlin2019bert} across 8 GPUs, and subsequently fine-tune them on the GLUE dev set.
%
The ANN baseline includes BERT \citep{devlin2019bert},  Q2BERT \citep{zhang2020ternarybert} and BiBERT \citep{qin2022bibert}.
Q2BERT employs 2-bit weights and 8-bit activations. BiBERT is the Binary Neural Network with 1-bit weights and activations, which was obtained by the Direction-Matching Distillation (DMD).
WE-Spikingformer$^\dagger$ (WE-Spikingformer with the T-LIF neuron) achieves an average accuracy of 66.3$\%$. Replacing T-LIF with the NI-LIF neuron yields comparable performance while significantly reducing pretraining time. These results demonstrate that both T-LIF and NI-LIF neurons are viable for language modeling. Moreover, the I-LIF neuron delivers similar accuracy with even faster spiking simulation. 
The experimental results are presented in Table~\ref{tab_glue}. All models are maintained at the same parameter scale as BERT$_\texttt{base}$ (109M). 
As a spike-driven, softmax-free transformer, WE-Spikingformer$^\dagger$ significantly outperforms other spike-based methods (LIF-BERT, PSN-BERT, and SpikeBERT). WE-Spikingformer$^\dagger$ vs. SpikeBERT vs. LIF-BERT. Acc: 66.3$\%$ vs. 59.7$\%$ vs. 34.6$\%$. In comparison, SpikeLM is a softmax-based spiking transformer and retains non-spiking activation GeLU \citep{hendrycks2016gaussian} in MLP blocks. These factors led to its relatively higher performance. 


\begin{table*}[htb]
\begin{center}
\begin{small}
\centering
\setlength{\tabcolsep}{0.9mm}
\fontsize{9.3pt}{\baselineskip}\selectfont
\caption{The results on Question Answering Tasks (QAT). "Avg." denotes "Average Accuracy". "E (mJ)" means "Energy consumption (mJ)". "T" means time step. "SD" denotes "Spike-Driven". }
\label{QA_tasks}
\begin{tabular}{lcp{1.7cm}<{\centering}p{0.8cm}
<{\centering}|ccccc|p{1.9cm}<{\centering}}
\toprule
{Model} & SD & E (mJ) & {T}  & {ARC-e} & {ARC-c} & {BoolQ} & {HeadQA}  & {OBQA} & {Avg. (\%)} \\ 
\midrule
SpikeLLM-7B \citep{xing2024spikellm} & \ding{55} & -  & 4 & 31.3 & 23.6   & 53.8 & -  & -  & -  \\
Qwen-1.5B \citep{qwen2} & \ding{55} & 3398.3  & - & 26.2 & 26.9   & 60.3 & 25.7  & 27.1 & 33.2  \\
\midrule
{WD-Spikingformer-0.4B}   & \ding{51}      & 238.4    & 4    &30.0  & 22.3 & 37.8 & 26.0 & 26.1 & {28.4} \\
\bottomrule
\end{tabular}
\end{small}
\end{center}
\end{table*}

\begin{table*}[htb]
\begin{center}
\begin{small}
\centering
\setlength{\tabcolsep}{1.8mm}
\caption{The results on Commonsense Reasoning Tasks (CRT). "Avg." denotes "Average Accuracy". "E (mJ)" means "Energy consumption (mJ)". "T" means time step. "SD" denotes "Spike-Driven".}
\label{commonsense_reasoning_tasks}
\fontsize{9.3pt}{\baselineskip}\selectfont
\begin{tabular}{lcp{1.7cm}<{\centering}p{0.7cm}<{\centering}|p{1.3cm}<{\centering}p{1.3cm}<{\centering}p{1.3cm}<{\centering}|p{1.8cm}<{\centering}}
\toprule
{Model} & SD & E (mJ) & {T}  & {HellaSwag} & {PIQA} & {Winograd} & {Avg. (\%)}\\ 
\midrule
SpikeLLM-7B \cite{xing2024spikellm} & \ding{55} & -  & 4 & 33.9 & 53.4 & 51.5  & 46.3   \\
Qwen-1.5B \citep{qwen2} & \ding{55} & 3398.3  & - & 26.5 & 53.7 & 52.8  & 44.3  \\
\midrule
{WD-Spikingformer-0.4B}  & \ding{51}  & 238.4    & 4   & 25.9 & 53.4 & 50.2 & {43.2} \\
\bottomrule
\end{tabular}
\end{small}
\end{center}
\end{table*}

\begin{table}[!tbp]
\centering
\caption{Ablation study for HardWTA on GLUE, QAT, and CRT. 
$\mathbb{M}(.)$ means causal mask for decoder-only spiking transformers. That is, $\mathbb{M}(\operatorname{Q}\operatorname{K}^\mathrm{T}*s)\operatorname{V}$ is the $\operatorname{Q}\operatorname{K}^\mathrm{T}\operatorname{V}*s$ in causal mask version.
Note that "\ding{55}" means the model does not converge in pretraining. "T" means time step. "Avg." denotes "Average Accuracy ($\%$)". 
}
\label{ablation_study_1}%
\begin{tabular}{lp{0.2cm}<{\centering}p{4.2cm}<{\centering}|p{1.1cm}<{\centering}}
\toprule
\multicolumn{1}{l}{Datasets} & T & Method &  Avg.  \\
\midrule
{\multirow{3}{*}{{GLUE}}} & 4 & $\operatorname{Q}\operatorname{K}^\mathrm{T}\operatorname{V}*s$  & 55.4  \\
 & 4 & $\operatorname{Softmax} (\operatorname{Q}\operatorname{K}^\mathrm{T}*s)\operatorname{V}$  & 66.8  \\
 & 4 & $\operatorname{Hard WTA}(\operatorname{Q}\operatorname{K}^\mathrm{T}*s)\operatorname{V}$ & 66.3  \\
\cmidrule(lr){1-4}
{\multirow{3}{*}{{QAT }}} & 4 & $\mathbb{M}(\operatorname{Q}\operatorname{K}^\mathrm{T}*s)\operatorname{V}$  & \ding{55}  \\
 & 4 & $\operatorname{Softmax} (\mathbb{M}(\operatorname{Q}\operatorname{K}^\mathrm{T}*s))\operatorname{V}$  & 28.7  \\
 & 4 & $\operatorname{Hard WTA}(\mathbb{M}(\operatorname{Q}\operatorname{K}^\mathrm{T}*s))\operatorname{V}$ & 28.4  \\
\cmidrule(lr){1-4}
{\multirow{3}{*}{{CRT }}}& 4 & $\mathbb{M}(\operatorname{Q}\operatorname{K}^\mathrm{T}*s)\operatorname{V}$ & \ding{55} \\
 & 4 & $\operatorname{Softmax} (\mathbb{M}(\operatorname{Q}\operatorname{K}^\mathrm{T}*s))\operatorname{V}$  & 43.3 \\
 & 4 & $\operatorname{Hard WTA}(\mathbb{M}(\operatorname{Q}\operatorname{K}^\mathrm{T}*s))\operatorname{V}$ & 43.2 \\
\bottomrule
\end{tabular}%
\end{table}%

\subsection{Question Answering Tasks} \label{Question_Answering_Tasks}
For question answering, we use the FineWeb-Edu \citep{lozhkov2024fineweb-edu} dataset, a high-quality subset of the FineWeb corpus curated for factual and educational content, and sample 0.1B tokens from it to pretrain WD-Spikingformer on 8 GPUs. 
We evaluate the model performance of WD-Spikingformer on a diverse set of Question-Answering Tasks (QAT), including ARC-e \citep{clark2018think}, ARC-c \citep{clark2018think}, BoolQ \citep{clark2019boolq}, HeadQA \citep{vilares2019head}, and OpenBookQA (OBQA) \citep{mihaylov2018can}. These tasks measure the generalization and reasoning abilities without task-specific finetuning. 

The experimental results are shown in Table~\ref{QA_tasks}.  
On one hand, due to the unavailability of the open-source model file for SpikeLLM \citep{xing2024spikellm} and the absence of reported energy consumption and results on HeadQA and OBQA,
our comparative analysis is limited to the average accuracy on the ARC-e, ARC-c, and BoolQ benchmarks.
Despite being 17.5 times smaller (0.4B vs. 7B parameters), WD-Spikingformer-0.4B achieves a competitive accuracy of $30.0\%$, approaching the $36.2\%$ accuracy of the much larger SpikeLLM-7B under the (T=4, W2A8) configuration.
On the other hand, WD-Spikingformer-0.4B reduces energy consumption by an order of magnitude (238.4 mJ vs. 3398.3 mJ, only $7\%$ of the energy) while maintaining a small accuracy gap ($28.4\%$ vs. $33.2\%$) against Qwen-1.5B, which adopts the same pretraining setup as WD-Spikingformer. These results highlight WD-Spikingformer’s superior energy–accuracy trade-off. 


\subsection{Commonsense Reasoning Tasks}
In this experiment, we evaluate the model performance of WD-Spikingformer on a diverse set of Commonsense Reasoning Tasks (CRT), including the HellaSwag \citep{zellers2019hellaswag}, PIQA \citep{bisk2020piqa}, and Winograd \citep{sakaguchi2021winogrande} datasets. The pretraining process is the same as shown in Section~\ref{Question_Answering_Tasks}. 

The experimental results are shown in Table~\ref{commonsense_reasoning_tasks}. 
Compared with SpikingLLM-7B and Qwen-1.5B, WD-Spikingformer-0.4B uses smaller parameters (0.4B vs. 1.5B and 7B) while its accuracy remains close to the larger models ($43.2\%$ vs. $44.3\%$ and $46.3\%$). 
These results highlight that WD-Spikingformer-0.4B delivers competitive performance despite its much smaller size.

\subsection{Ablation Study}
In this part, we carry out the ablation study for HardWTA, different WTA methods, and time steps.

\textbf{Ablation study for HardWTA.} 
We carry out the ablation study for HardWTA on three tasks: GLUE datasets, QAT, and CRT. We compare HardWTA with the vision-oriented spiking self-attention \citep{zhou2023spikformer} ($\operatorname{Q}\operatorname{K}^\mathrm{T}\operatorname{V}*s$) and softmax-based version ($\operatorname{Softmax} (\operatorname{Q}\operatorname{K}^\mathrm{T}*s)\operatorname{V}$) \citep{xing2024spikelm} in these language tasks. 



The experimental results are shown in Table~\ref{ablation_study_1}, and indicate that: 
1) Vision-oriented spiking self-attention ($\operatorname{Q}\operatorname{K}^\mathrm{T}\operatorname{V}*s$) does not work well in language modeling, often leading to convergence difficulties or suboptimal performance. $\operatorname{Hard WTA}(\operatorname{Q}\operatorname{K}^\mathrm{T}*s)\operatorname{V}$ vs. ($\operatorname{Q}\operatorname{K}^\mathrm{T}\operatorname{V}*s$). GLUE: 66.3$\%$ vs. 55.4$\%$; QAT: 28.4$\%$ vs. Not converging in pretraining; CRT: 43.2$\%$ vs. Not converging in pretraining.
2) The Winner-Take-All layer can approximately replace the softmax layer in self-attention computation of spiking transformers in language modeling while enjoying high energy efficiency during inference. $\operatorname{Hard WTA}(\operatorname{Q}\operatorname{K}^\mathrm{T}*s)\operatorname{V}$ vs. $\operatorname{Softmax} (\operatorname{Q}\operatorname{K}^\mathrm{T}*s)\operatorname{V}$. GLUE: 66.3$\%$ vs. 66.8$\%$; QAT: 28.4$\%$ vs. 28.7$\%$; CRT: 43.2$\%$ Vs. 43.3$\%$. 
The three tasks ablation study verified the effectiveness of HardWTA, tailor-made for the spiking transformers in language modeling.

\textbf{Ablation study for different WTAs and time steps.}
We carry out the ablation study for different WTAs and time steps on GLUE datasets. We set the hyperparameters $\operatorname{K}=3$  in $\operatorname{Top-K WTA}(\operatorname{Q}\operatorname{K}^\mathrm{T}*s)\operatorname{V}$. The effects of the K hyperparameters are shown in the Appendix \ref{Topk_on_glue}.

The experimental results are shown in Table~\ref{ablation_study_2}, which indicates that the performance of Hard WTA, Top-K WTA, and Sparsemax is very similar on GLUE (65.7$\%$ vs. 66.3$\%$ vs. 66.6$\%$), making Hard WTA the more cost-effective option. When the time step is increased to 8, the performance of WE-Spikingformer further improves to an average accuracy of 67.4$\%$.


\begin{table}[!tbp]
\centering
\caption{Ablation study on different WTAs and time steps on GLUE datasets. "T" means time step. "Avg." denotes "Average Accuracy ($\%$)". 
}
\label{ablation_study_2}%
\begin{tabular}{lp{0.2cm}<{\centering}p{4.2cm}<{\centering}|p{1.1cm}<{\centering}}
\toprule
\multicolumn{1}{l}{Datasets} & T & Method &  Avg.  \\
\midrule
\cmidrule(lr){1-4}
{\multirow{6}{*}{{GLUE }}} & 4 & $\operatorname{Hard WTA}(\operatorname{Q}\operatorname{K}^\mathrm{T}*s)\operatorname{V}$ & 65.7 \\
 & 4 & $\operatorname{Top-K WTA}(\operatorname{Q}\operatorname{K}^\mathrm{T}*s)\operatorname{V}$  & 66.3 \\
 & 4 & $\operatorname{Sparsemax}(\operatorname{Q}\operatorname{K}^\mathrm{T}*s)\operatorname{V}$ & 66.6 \\
 \cmidrule(lr){2-4}

 & 8 & $\operatorname{Hard WTA}(\operatorname{Q}\operatorname{K}^\mathrm{T}*s)\operatorname{V}$ & 67.4 \\
 & 4 & $\operatorname{Hard WTA}(\operatorname{Q}\operatorname{K}^\mathrm{T}*s)\operatorname{V}$ & 65.7 \\
 & 2 & $\operatorname{Hard WTA}(\operatorname{Q}\operatorname{K}^\mathrm{T}*s)\operatorname{V}$ & 62.3 \\
\bottomrule
\end{tabular}%
\end{table}%

\subsection{Discussion} \label{discussion}
In this part, we discuss the scaling characteristics of our Winner-Take-All spiking transformer from the model parameter and pretraining data perspectives.

\begin{figure}[!t]
\centering
\includegraphics[width=0.48\textwidth]{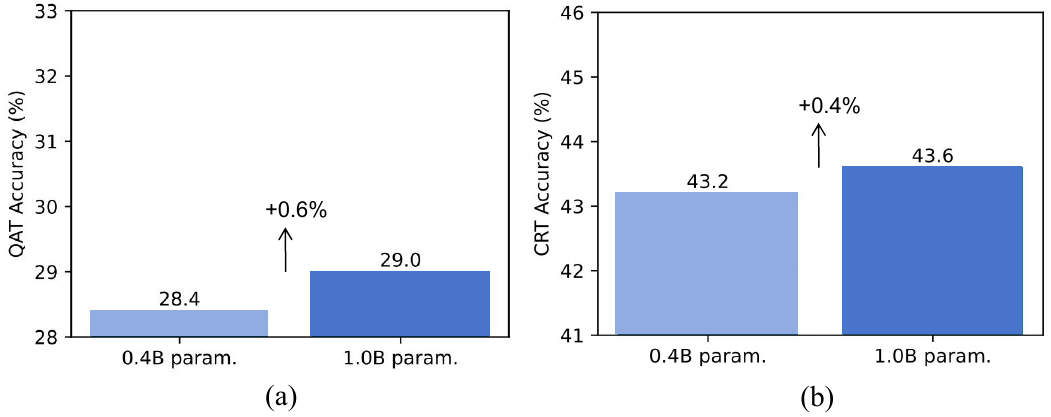}
\caption{Increasing model Parameters for WE-Spikingformer pretraining on (a) Question-Answering Tasks (QAT), and (b) Commonsense Reasoning Tasks (CRT).}
\label{discussion_0}
\end{figure}

\textbf{Parameter scaling.} The experimental results are shown in Figure \ref{discussion_0}. We increase the model parameters of WD-Spikingformer from 0.4B to 1.0B parameters. The experimental results show that the performance of WD-Spikingformer can be further improved on both QAT and CRT. WE-Spikingformer-0.4B vs. WE-Spikingformer-1.0B. QAT: 28.4$\%$ vs. 29.0$\%$; CRT: 43.2$\%$ vs. 43.6$\%$.

\textbf{Pretraining data scaling.}
The experimental results are presented in Figure.\ref{discussion}. In this part, we increase the tokens for WE-Spikingformer-1.0B pretraining and validated it on QAT and CRT. 
We found that WE-Spikingformer-1.0B can achieve a further significant improvement on both QAT and CRT. WE-Spikingformer-1.0B (0.1Btokens) vs. WE-Spikingformer-1.0B (0.5B tokens). QAT: 29.0$\%$ vs. 31.9$\%$; CRT: 43.6$\%$ vs. 44.1$\%$.
In particular, improving the pretraining data of WE-Spikingformer-1.0B from 0.1B tokens to 0.5B tokens, the performance of WE-Spikingformer-1.0B achieves a significant performance improvement of 1.9$\%$ on QAT, demonstrating that WE-Spikingformer continues to significantly benefit from larger-scale pretraining.

In summary, we verified that our model's performance can be further improved by increasing the parameter scale or pretraining data scale, validating the potential scalability of the Winner-Take-All spiking transformer.

\begin{figure}[!t]
\centering
\includegraphics[width=0.48\textwidth]{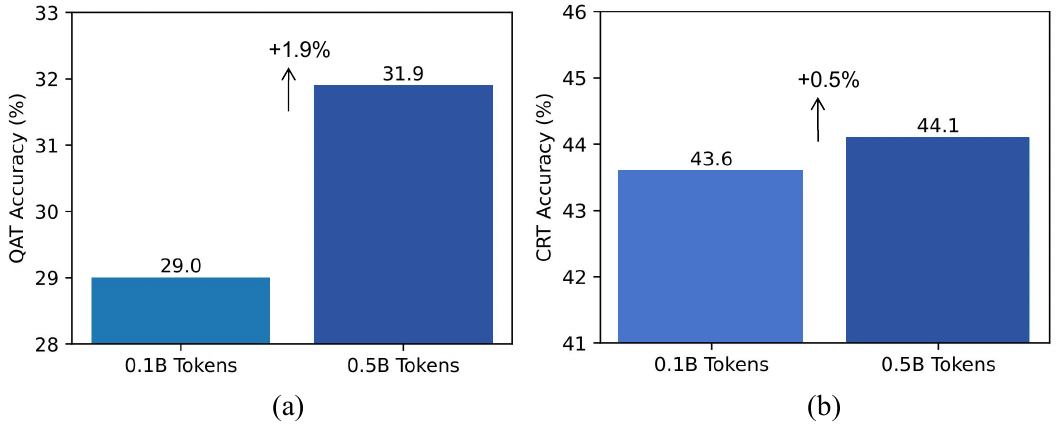}
\caption{Increasing tokens for WE-Spikingformer-1.0B pretraining on (a) Question-Answering Tasks (QAT), and (b) Commonsense Reasoning Tasks (CRT).}
\label{discussion}
\end{figure}

\section{Conclusion}
In this work, we explore softmax-free spiking transformers for language modeling by incorporating Winner-Take-All (WTA) mechanisms into spike-driven self-attention. We propose two novel attention modules, namely WTA Spiking Self-Attention (WSSA) and Causal WTA Spiking Self-Attention(CWSSA).
Based on these modules, we design WE-Spikingformer for masked language modeling and WD-Spikingformer for causal language modeling.
%
Our approach systematically extends directly trained spiking transformers from vision to language. Extensive experiments across 16 datasets spanning natural language understanding, question answering tasks, and commonsense reasoning tasks demonstrate the effectiveness of our models. These results underscore the potential of Spiking Transformers as a foundation for biologically inspired, energy-efficient, and general-purpose language modeling.

\textbf{Limitation:} Our energy consumption estimates are based on theoretical calculations and do not include measurements on real neuromorphic hardware. In addition, with its current scale parameters, WD-Spikingformer is not yet capable of effectively handling large-scale, long-context generative downstream tasks. The above is reserved for future work.









\bibliography{WTA}
\bibliographystyle{icml2026}

\newpage
\appendix
\onecolumn
\section*{Appendix}
\section{Version Statement}
We have further refined and improved the paper, building upon the previous version available at \url{https://openreview.net/forum?id=7PKGMNcM0w}.

\section{Dataset Introduction}
\textbf{General Language Understanding Evaluation (GLUE).} GLUE benchmark \citep{wang2018glue} is a widely adopted framework for assessing natural language understanding (NLU) models across a variety of tasks, including single-sentence classification, sentence-pair classification, and linguistic acceptability. 
In our experiments, we focus on eight representative datasets within GLUE:
MNLI (Multi-Genre Natural Language Inference): Predict whether a hypothesis is entailed, contradicted, or neutral with respect to a given premise across multiple text genres.
QQP (Quora Question Pairs): Detect whether two questions from Quora convey the same meaning.
QNLI (Question Natural Language Inference): Reformulated from QA, determines whether a sentence contains the answer to a question.
SST-2 (Stanford Sentiment Treebank): Perform binary sentiment classification on movie reviews.
CoLA (Corpus of Linguistic Acceptability): Judge whether a sentence is grammatically acceptable.
STS-B (Semantic Textual Similarity Benchmark): Measure sentence-level semantic similarity on a scale from 0 to 5.
MRPC (Microsoft Research Paraphrase Corpus): Identify whether two sentences are paraphrases.
RTE (Recognizing Textual Entailment): Decide whether a hypothesis can be inferred from a premise, based on multiple entailment datasets.

\textbf{Question Answering Tasks (QAT).}
ARC-e \citep{clark2018think} and ARC-c (AI2 Reasoning Challenge, easy and challenge subsets) assess scientific knowledge and reasoning skills, with ARC-e focusing on simpler multiple-choice science questions and ARC-c including more complex ones that require advanced reasoning. 
BoolQ (Boolean Questions) \citep{clark2019boolq} is a yes/no question-answering dataset derived from natural queries, requiring models to determine the truthfulness of a statement given a supporting passage. 
HeadQA (Head Question Answering) \citep{vilares2019head} is a multilingual medical question-answering benchmark composed of exams for healthcare professionals, testing domain-specific knowledge. 
OBQA (OpenBookQA) \citep{mihaylov2018can} evaluates a model’s ability to answer elementary science questions by combining provided core facts with external common knowledge.

\textbf{Commonsense Reasoning Tasks (CRT).}
HellaSwag \citep{zellers2019hellaswag} is a large-scale dataset for grounded commonsense inference, where models must select the most plausible continuation of a given context. 
PIQA (Physical Interaction Question Answering) \citep{bisk2020piqa} focuses on physical commonsense reasoning, requiring models to choose the more feasible solution to everyday tasks. 
Winograd (Winograd Schema Challenge) \citep{sakaguchi2021winogrande} is a coreference resolution benchmark designed to test commonsense reasoning by requiring models to resolve pronoun references that cannot be disambiguated by syntax alone.

\section{Energy consumption}
SNNs replace traditional multiply-accumulate (MAC) operations with low-power accumulate (AC) operations. For ANNs, the overall energy consumption can be directly evaluated by their MACs. For example, given a linear layer with input dimension $m$ and output dimension $n$, its energy consumption can be calculated by:
\begin{gather}
E^{\operatorname{Ann}}_{\operatorname{Linear}}=m \times n \times E_{M A C} ,
\label{eq:energy_ann}
\end{gather}
SNNs convert MAC-based matrix multiplications to pure accumulate operations (ACs). For SNNs, given the same example as the ANN case, the theoretical energy consumption of the linear layer can be calculated by:
\begin{gather}
E^{\operatorname{Snn}}_{\operatorname{Linear}}=m \times n \times E_{A C} \times f r \times T ,
\label{eq:energy_snn}
\end{gather}
where $fr$ is the firing rate of the layer, and $T$ is the simulation time step of the spiking neuron. Refer to previous works\citep{kundu2021hire,zhou2023spikformer,zhou2023spikingformer,panda2020toward,yao2023attention}. 
We assume that the MAC and AC operations are implemented on the 45nm hardware \citep{horowitz20141}, where $E_{MAC}=4.6pJ$ and $E_{AC}=0.9pJ$.
The energy consumption of the models in the paper is calculated by reasoning about 512 tokens.

\section{Vision-oriented Spiking Self-Attention}
Vision-oriented Spiking Self-Attention \citep{zhou2023spikformer, zhou2023spikingformer} use a sparse spike-form $\mathbf{Q, K, V}$ for vision task modeling without softmax operation and floating-point matrix multiplication. The calculation process of vision-oriented spiking self-attention is formulated as follows:
\begin{align} \label{SSA}
\textbf{X}^{\prime} &= \operatorname{SN}(\textbf{X}), \\
\textbf{Q} &= \operatorname{SN}_{Q}(\operatorname{Linear}_{Q}(\textbf{X}^{\prime})), \\
\textbf{K} &= \operatorname{SN}_{K}(\operatorname{Linear}_{K}(\textbf{X}^{\prime})), \\
\textbf{V} &= \operatorname{SN}_{V}(\operatorname{Linear}_{V}(\textbf{X}^{\prime})), \\
\operatorname{SSA}^{\prime}(\mathbf{Q}, \mathbf{K}, \mathbf{V})&=\mathrm{S N}\left(\mathbf{Q} \mathbf{K}^{\mathrm{T}} \mathbf{V} * s\right),
\end{align} 
where $\mathbf{Q}, \mathbf{K}, \mathbf{V} \in \mathcal{R}^{T \times N \times D}$,
the spike-form $\mathbf{Q}, \mathbf{K}, \mathbf{V}$ are computed by learnable linear layers. $s$ is a scaling factor. $\mathrm{SN}$ means spiking neuron layer.
The calculation of SSA avoids floating-point multiplication, meeting the property of SNNs.

\section{Hard Winner-Take-All in SNN} \label{Hard_WTA}
Given an input vector $\mathbf{x}=\left[x_1, x_2, \ldots, x_n\right]$, the softmax function is defined as:
\begin{equation}
\operatorname{softmax}\left(x_i\right)=\frac{e^{x_i}}{\sum_{j=1}^n e^{x_j}}
\end{equation}

Introduce the temperature parameter $\tau > 0$, the temperature parameterization softmax operation can be described as follows:
\begin{equation}
\operatorname{softmax}_\tau\left(x_i\right)=\frac{e^{x_i / \tau}}{\sum_{j=1}^n e^{x_j / \tau}}
\end{equation}
When $\tau = 1$, the temperature parameterization softmax operation is equal to the standard softmax.

\begin{lemma}
  \label{lem:usefullemma}
    As $\tau \rightarrow 0^{+}$, the temperature parameterization softmax converges to Hard WTA in SNN.
$$
\lim _{\tau \rightarrow 0^{+}} \text{softmax}_\tau(x_i) = \begin{cases}1 & \text { if } i=\arg \max _j u_j \\ 0 & \text { otherwise }\end{cases}
$$
\end{lemma}

Assume $x_{max}=max(x_{j})$, and we assume that $x_{1}=x_{max}$
\begin{equation}
\operatorname{softmax}_\tau\left(x_1\right)=\frac{e^{x_1 / \tau}}{e^{x_1 / \tau}+\sum_{j=2}^n e^{x_j / \tau}}=\frac{1}{1+\sum_{j=2}^n e^{\left(x_j-x_1\right) / \tau}}
\end{equation}

For all $j\ne1$, $x_j-x_1 <0$, when $T \rightarrow 0^{+}$:
\begin{equation}
e^{\left(x_j-x_1\right) / \tau} \rightarrow 0, \quad \forall j \neq 1
\end{equation}

Then:
\begin{equation}
\lim _{\tau \rightarrow 0^{+}} \operatorname{softmax}_\tau\left(x_1\right)=\frac{1}{1+0}=1
\end{equation}

Similarly:
\begin{equation}
\lim _{\tau \rightarrow 0^{+}} \operatorname{softmax}_\tau\left(x_i\right)=\lim _{\tau \rightarrow 0^{+}} \frac{e^{\left(x_i-x_1\right) / \tau}}{1+\sum_{j=2}^n e^{\left(x_j-x_1\right) / \tau}}=0
\end{equation}

Therefore, Hard WTA in SNN is an extremely sparse softmax with one-hot encoding.

\section{Sparsemax} \label{appendix_sparsemax}

As shown in \citet{martins2016softmax}, Sparsemax is defined as:
\begin{equation}\label{eq:sparsemax}
\operatorname{sparsemax}(\boldsymbol{z}):=\underset{\boldsymbol{p} \in \Delta^{K-1}}{\operatorname{argmin}}\|\boldsymbol{p}-\boldsymbol{a}\|^2
\end{equation}
Sparsemax computes the Euclidean projection of the input vector $\boldsymbol{a}$ onto the probability simplex. This projection often lies on the boundary of the simplex, resulting in sparse outputs. 
The solution of Eq.~\ref{eq:sparsemax} is shown in Eq.~\ref{Sparsemax} ($y_i=\max \left\{a_i-\tau, 0\right\}, \quad i=1, \ldots, N .$), where $\tau$ is the adaptive threshold. Furthermore, $\tau$  can be expressed as follows. 
Let $a_{(1)} \ge a_{(2)} \ge \ldots \ge a_{(K)}$ be the sorted coordinates of $\boldsymbol{a}$, and define 
$k(\boldsymbol{a}) := \max\left\{k \in [K]\,\,|\,\, 1+ka_{(k)} > \sum_{j\le k} a_{(j)}\right\}$.
Then,
\begin{eqnarray}\label{eq:threshold_closedform}
\tau(\boldsymbol{a}) = \frac{\left(\sum_{j\le k(\boldsymbol{a})} a_{(j)}\right) - 1}{k(\boldsymbol{a})}
= 
\frac{\left(\sum_{j \in S(\boldsymbol{a})} a_j\right) - 1}{|S(\boldsymbol{a})|},
\end{eqnarray}
where $S(\boldsymbol{a}) := \{j \in [K]\,\,|\,\, \mathrm{sparsemax}_j(\boldsymbol{a}) > 0\}$ is the support of 
$\mathrm{sparsemax}(\boldsymbol{a})$.
In essence, all we need for evaluating the Sparsemax transformation is to compute the threshold $\tau(\boldsymbol{a})$; 
all coordinates above this threshold (the ones in the set $S(\boldsymbol{a})$) will be shifted by this amount, 
and the others will be truncated to zero. Therefore, Sparsemax can also be seen as an adaptive threshold form for the Top-K Winner-Take-All.

\section{Detailed Results}
This part provides more detailed results for Section.\ref{discussion}, which is shown in Table.\ref{detail_qat} and Table.\ref{detail_crt}. 
In particular, when scaled up to 1.0B parameters, WD-Spikingformer further improves its accuracy, nearly matching that of Qwen-1.5B (43.6$\%$ vs. 44.3$\%$). However, its energy consumption is only about 17$\%$ of that of the Qwen-1.5B (574.7 mJ vs. 3398.3 mJ).

\begin{table*}[htb]
\begin{center}
\begin{small}
\centering
\setlength{\tabcolsep}{0.9mm}
\fontsize{9.3pt}{\baselineskip}\selectfont
\caption{The results on Question Answering Tasks (QAT). "Avg." denotes "Average Accuracy". "E (mJ)" means "Energy consumption (mJ)". "T" means time step. "SD" denotes "Spike-Driven". WD-Spikingformer-1.0B$^{*}$ means WD-Spikingformer-1.0B with 0.5B tokens during pretraining.}
\label{detail_qat}
\begin{tabular}{lcp{1.7cm}<{\centering}p{0.8cm}
<{\centering}|ccccc|p{1.9cm}<{\centering}}
\toprule
{Model} & SD & E (mJ) & {T}  & {ARC-e} & {ARC-c} & {BoolQ} & {HeadQA}  & {OBQA} & {Avg. (\%)} \\ 
\midrule
SpikeLLM-7B \citep{xing2024spikellm} & \ding{55} & -  & 4 & 31.3 & 23.6   & 53.8 & -  & -  & -  \\
Qwen-1.5B \citep{qwen2} & \ding{55} & 3398.3  & - & 26.2 & 26.9   & 60.3 & 25.7  & 27.1 & 33.2  \\
\midrule
{WD-Spikingformer-0.4B}   & \ding{51}      & 238.4    & 4    &30.0  & 22.3 & 37.8 & 26.0 & 26.1 & {28.4} \\
{WD-Spikingformer-1.0B}   & \ding{51}      & 574.7   & 4    &33.8  & 20.3 & 37.8 & 26.4 & 26.6 & {29.0} \\
{WD-Spikingformer-1.0B$^{*}$}   & \ding{51}      & 574.7   & 4    &41.9 & 20.4 &42.1 & 26.7 & 28.2 & {31.9} \\
\bottomrule
\end{tabular}
\end{small}
\end{center}
\end{table*}

\begin{table*}[htb]
\begin{center}
\begin{small}
\centering
\setlength{\tabcolsep}{1.8mm}
\caption{The results on Commonsense Reasoning Tasks. "Avg." denotes "Average Accuracy". "E (mJ)" means "Energy consumption (mJ)". "T" means time step. "SD" denotes "Spike-Driven". WD-Spikingformer-1.0B$^{*}$ means WD-Spikingformer-1.0B with 0.5B tokens during pretraining.}
\label{detail_crt}
\fontsize{9.3pt}{\baselineskip}\selectfont
\begin{tabular}{lcp{1.7cm}<{\centering}p{0.7cm}<{\centering}|p{1.3cm}<{\centering}p{1.3cm}<{\centering}p{1.3cm}<{\centering}|p{1.8cm}<{\centering}}
\toprule
{Model} & SD & E (mJ) & {T}  & {HellaSwag} & {PIQA} & {Winograd} & {Avg. (\%)}\\ 
\midrule
SpikeLLM-7B \cite{xing2024spikellm} & \ding{55} & -  & 4 & 33.9 & 53.4 & 51.5  & 46.3   \\
Qwen-1.5B \citep{qwen2} & \ding{55} & 3398.3  & - & 26.5 & 53.7 & 52.8  & 44.3  \\
\midrule
{WD-Spikingformer-0.4B}  & \ding{51}  & 238.4    & 4   & 25.9 & 53.4 & 50.2 & {43.2} \\
{WD-Spikingformer-1.0B}  & \ding{51}  & 574.7    & 4   & 26.0 & 55.8 & 49.1 & {43.6} \\
{WD-Spikingformer-1.0B$^{*}$}  & \ding{51}  & 574.7    & 4   & 26.4 & 55.7 & 50.3 & {44.1} \\
\bottomrule
\end{tabular}
\end{small}
\end{center}
\end{table*}

\section{Detailed Results on Spiking Neuron Comparison}
For WE-Spikingformer pretraining, the training time comparison on 8 × 4090 GPUs is shown in Table \ref{training_time}. NI-LIF significantly boosts training simulation speed compared to T-LIF, while maintaining comparable model performance.

\begin{table}[!tbp]
\centering
\caption{Spiking neuron comparison for WE-Spikingformer pretraining.
}
\label{training_time}%
\begin{tabular}{lp{2.2cm}<{\centering}|p{1.cm}<{\centering}p{4.2cm}<{\centering}p{3.1cm}<{\centering}}
\toprule
{Backbone} & Neuron & T & Average accuracy ($\%$) &  Training time ( hours)  \\
\midrule
WE-Spikingformer & NI-LIF	& 4 & 65.7 & 26  \\
WE-Spikingformer & T-LIF	& 4 & 66.3 & 121  \\
\bottomrule
\end{tabular}%
\end{table}%

\section{Additional Ablation Study for Surrogate Gradient on GLUE.}
In previous experiments, we adopted softmax as the surrogate gradient for Hard WTA, motivated by our theoretical analysis in Appendix \ref{Hard_WTA}, where we prove that Hard WTA can be viewed as an extremely sparse softmax with one-hot outputs.
In this experiment, we further conduct an ablation study on the choice of surrogate gradient. Specifically, we replace softmax with ReLU \cite{agarap2018deep} as the surrogate gradient for Hard WTA. The experimental results on GLUE datasets are reported in the Table.\ref{ablation_sg}.

\begin{table*}[htb]
\begin{center}
\begin{small}
\centering
\caption{Ablation study for Surrogate Gradient on GLUE datasets. }
\label{ablation_sg}
\setlength{\tabcolsep}{1.7mm}
\fontsize{9.3pt}{\baselineskip}\selectfont
\begin{tabular}{lp{1.2cm}<{\centering}|cccccccc|p{1.6cm}<{\centering}}
\toprule
{Surrogate Gradient}  & {T}  & {MNLI\_{mm}} & {QQP} & {QNLI} & {SST-2} & {CoLA} & {STS-B} & {MRPC\_{F1}} & {RTE} & {Avg. (\%)} \\ 
\midrule
Softmax    & 4    & 73.4 & 83.8 & 78.3 & 85.6 & 23.7 & 56.0 & 76.4 & 48.4 & {65.7}  \\
ReLU    & 4    & 31.8 & 63.2 & 50.5 & 50.9 & 0 & 4.3 & 75.0 & 47.3 & {40.4}  \\
\bottomrule
\end{tabular}
\end{small}
\end{center}
\end{table*}

\section{Additional Ablation Study for Top-K Winner-Take-All on GLUE.} \label{Topk_on_glue}
In this section, we conduct an ablation study of the Top-K Winner-Take-All (WTA) mechanism on the GLUE benchmark to investigate the impact of the hyperparameter K on model performance. According to the Top-K WTA principle, increasing K reduces activation sparsity, thereby leading to lower energy efficiency. Based on this trade-off, we evaluate WE-Spikingformer with K=1, 3, and 5 on the GLUE datasets. The experimental results are summarized in Table~\ref{ablation_topk}.

\begin{table*}[htb]
\begin{center}
\begin{small}
\centering
\caption{Ablation study for Top-K Winner-Take-All on GLUE datasets. Top-K WTA (K=1) is equal to Hard WTA.}
\label{ablation_topk}
\setlength{\tabcolsep}{1.7mm}
\fontsize{9.3pt}{\baselineskip}\selectfont
\begin{tabular}{lp{1.2cm}<{\centering}|cccccccc|p{1.6cm}<{\centering}}
\toprule
{Method}  & {T}  & {MNLI\_{mm}} & {QQP} & {QNLI} & {SST-2} & {CoLA} & {STS-B} & {MRPC\_{F1}} & {RTE} & {Avg. (\%)} \\ 
\midrule
$\operatorname{Softmax} (\operatorname{Q}\operatorname{K}^\mathrm{T}*s)\operatorname{V}$           & 4    & 72.5 & 84.7 & 76.0 & 87.2 & 24.4 & 54.5 & 79.7 & 55.6 & 66.8 \\
\midrule
Top-K WTA (K=1)    & 4    & 73.4 & 83.8 & 78.3 & 85.6 & 23.7 & 56.0 & 76.4 & 48.4 & {65.7}  \\
Top-K WTA (K=3)    & 4    & 73.4 & 84.0 & 78.8 & 86.1 & 24.1 & 55.9 & 78.2 & 49.8 & {66.3}  \\
Top-K WTA (K=5)    & 4    & 73.1 & 84.4 & 79.4 & 86.0 & 24.8 & 55.4 & 79.1 & 50.1 & {66.5}  \\
\bottomrule
\end{tabular}
\end{small}
\end{center}
\end{table*}

As the sequence length increases, the computational energy consumption of both Softmax and Top-K WTA rises; however, Top-K WTA still maintains an advantage. For the same sequence length,
Softmax involves exponentiation, summation, and division. At the hardware level, this requires exponentiation units, adders, and multipliers, with exponentiation and global normalization (division) being particularly computationally expensive\cite{zhou2023spikformer,zhou2023spikingformer}.
In our Spiking Transformer for language modeling, we replace the Softmax operation with a winner-take-all (WTA) mechanism. Specifically, Top-K WTA can be efficiently implemented using comparator-based hardware, such as comparator trees or heap-based selection logic. Unlike Softmax, which relies on floating-point exponentiation and global normalization, WTA primarily uses comparison and bit-masking operations, thereby avoiding these computationally expensive steps. \cite{10931119}
In addition, \citet{10931119} reports that this Top-K mechanism achieves approximately an 8× improvement in energy efficiency over the standard Softmax layer.

\section{Additional Ablation Study for different WTAs and time steps on CRT.}

\begin{table}[htbp]
\centering
\caption{Ablation study for different WTAs and time steps on CRT. 
}
\label{ablation_study_3}%
\begin{tabular}{lp{2.2cm}<{\centering}p{4.2cm}<{\centering}|p{4.0cm}<{\centering}}
\toprule
\multicolumn{1}{l}{Datasets} & Time step & Method &  Average Accuracy ($\%$)  \\
\midrule
\cmidrule(lr){1-4}
{\multirow{6}{*}{{CRT }}} & 4 & $\operatorname{Hard WTA}(\mathbb{M}(\operatorname{Q}\operatorname{K}^\mathrm{T}*s))\operatorname{V}$ & 43.2 \\
 & 4 & $\operatorname{Top-K WTA}(\mathbb{M}(\operatorname{Q}\operatorname{K}^\mathrm{T}*s))\operatorname{V}$  & 42.9 \\
 & 4 & $\operatorname{Sparsemax}(\mathbb{M}(\operatorname{Q}\operatorname{K}^\mathrm{T}*s))\operatorname{V}$ & 43.2 \\
 \cmidrule(lr){2-4}

 & 8 & $\operatorname{Hard WTA}(\mathbb{M}(\operatorname{Q}\operatorname{K}^\mathrm{T}*s))\operatorname{V}$ & 43.7 \\
 & 4 & $\operatorname{Hard WTA}(\mathbb{M}(\operatorname{Q}\operatorname{K}^\mathrm{T}*s))\operatorname{V}$ & 43.2 \\
 & 2 & $\operatorname{Hard WTA}(\mathbb{M}(\operatorname{Q}\operatorname{K}^\mathrm{T}*s))\operatorname{V}$ & 40.1 \\
\bottomrule
\end{tabular}%
\end{table}%

Table \ref{ablation_study_3} shows the ablation study for different WTAs and time steps on CRT. The performance of Hard WTA, Top-K WTA (K=3), and Sparsemax is very similar on CRT (43.2$\%$ vs. 42.9$\%$ vs. 43.2$\%$), making Hard WTA the more cost-effective option. When the time step is increased to 8, the performance of the WTA-based decoder-only spiking transformer (WD-Spikingformer-0.4B) further improves to an average accuracy of 43.7$\%$.

\section{Differences between HardWTA and MaxPooling}
In this section, we highlight the key differences between the HardWTA mechanism and max pooling:

\textbf{Differences in information retention:} MaxPooling retains the maximum value within a local receptive field. In contrast, Hard WTA assigns a value of 1 to the neuron with the highest response while suppressing all others to 0, yielding a binary and highly sparse representation. Here is a simple example illustrating the operation process: Input: [0.2, 0.8, 0.5, 0.1], MaxPooling: [0.8], HardWTA: [0, 1, 0, 0].

\textbf{Differences in dimensionality changes:} MaxPooling reduces the spatial resolution of the feature map (e.g., "28×28 → 14×14"). In contrast, Hard WTA preserves the original dimensionality, performing competitive selection without altering the feature size.

\textbf{Differences in scope of operation:} MaxPooling operates over local spatial neighborhoods within each feature map (i.e., along the "H × W" dimensions). By contrast, Hard WTA is typically applied across the channel dimension, enforcing competition among neurons at the same spatial location.

Specifically, 
MaxFormer applies MaxPooling within the patch embedding module in vision tasks. Beyond downsampling, this operation enhances the representation of high-frequency components, facilitating more effective visual feature extraction.
SpikePool incorporates MaxPooling within a $\{$LIF–MaxPool–Conv–BN$\}$ block to replace the whole spiking self-attention mechanism, targeting small-scale event-based data.
Ours employs Winner-Take-All mechanism to replace the Softmax operation in self-attention, enabling sparse, spike-driven competition for spike-based language modeling.

\end{document}